\documentclass[11pt]{article}

\usepackage[preprint]{acl}

\usepackage{times}
\usepackage{latexsym}
\usepackage{forest}
\usepackage{expex}
\usepackage{pgfplots}
\pgfplotsset{compat=1.18}
\usepackage{array}
\usepackage{amsmath} 
\usepackage{graphicx}
\usepackage{subcaption} 
\usepackage{pgfplots}
\pgfplotsset{compat=1.18}
\usepgfplotslibrary{groupplots}
\usepackage{pgfplots}
\pgfplotsset{compat=1.18} 
\usepackage{xcolor}
\usepackage{amsmath}
\usepackage{booktabs}
\usepackage{array}
\usepackage{pgfplots}
\pgfplotsset{compat=1.18}
\usepackage{xcolor}
\usepackage{pgfplots}
\pgfplotsset{compat=1.18}
\usepackage{graphicx}
\usepackage{xcolor}
\usepackage{tikz}
\usetikzlibrary{arrows.meta,positioning,calc}

\usepackage{pgfplots}
\pgfplotsset{compat=1.18}
\usepgfplotslibrary{groupplots}

\definecolor{mplblue}{RGB}{31,119,180}
\definecolor{mplorange}{RGB}{255,127,14}
\definecolor{mplgreen}{RGB}{44,160,44}
\definecolor{mplpink}{RGB}{227,119,194}

\usepackage{fvextra}
\DefineVerbatimEnvironment{prompt}{Verbatim}{
  breaklines=true,
  breakanywhere=true,
  fontsize=\small
}

\lingset{
  aboveexskip=0.6ex,
  belowexskip=0.6ex,
  interpartskip=0.2ex,
  glstyle=nlevel,
  glhangstyle=none,
  everygla=\itshape,  
  everyglb=\upshape,  
}

\usepackage[T1]{fontenc}

\usepackage[utf8]{inputenc}

\usepackage{microtype}

\usepackage{inconsolata}

\usepackage{graphicx}

\title{Clause-Internal or Clause-External? Testing Turkish Reflexive Binding in Adapted versus Chain-of-Thought Large Language Models}

\author{
  Sercan Karaka\c{s} \\
  University of Chicago \\
  \texttt{skarakas@uchicago.edu}
}

\begin{document}
\maketitle
\begin{abstract}
This study evaluates whether state-of-the-art large language models capture the binding relations of Turkish reflexive pronouns. We construct a balanced evaluation set of 100 Turkish sentences that systematically pit local against non-local antecedents for the reflexives \textit{kendi} and \textit{kendisi}. We compare two contrasting systems: an OpenAI chain-of-thought model optimized for multi-step reasoning and Trendyol-LLM-7B-base-v0.1, a \textsc{LLaMA}-2--derived model extensively fine-tuned on Turkish data. Antecedent choice is assessed using a combined paradigm that integrates sentence-level perplexity with a forced-choice comparison between minimally differing continuations. Overall, Trendyol-LLM favors local bindings in approximately 70\% of trials, exhibiting a robust locality bias consistent with a preference for structurally proximate antecedents. By contrast, the OpenAI model (o1 Mini) distributes its choices nearly evenly between local and long-distance readings, suggesting weaker or less consistent sensitivity to locality in this binding configuration. Taken together, these results reveal a marked contrast in binding behavior across the two systems and motivate closer analysis of how model architecture, training data, and inference-time reasoning strategies shape the representation of Turkish anaphoric dependencies.

\end{abstract}


\section{Introduction}
Recent advances in Large Language Models (LLMs) have yielded systems that achieve state-of-the-art performance across a diverse array of tasks, including open-domain question answering, text generation, summarization, and creative writing (e.g., \citealp{openai2023gpt4,dubey2024llama3,anil2023palm2,riviere2024gemma2}). Beyond their well-documented linguistic versatility, these models appear capable of tackling sophisticated problem-solving challenges in domains such as software engineering and symbolic mathematics. Yet a central question remains unresolved: do contemporary LLMs truly ``reason'' in a human-like, logically compositional sense, or do they succeed mainly by exploiting vast training corpora to match surface-level statistical patterns? A growing body of evidence suggests the latter explanation still dominates. Diagnostic studies argue that LLMs often default to probabilistic pattern matching rather than rule-governed logical inference, revealing systematic blind spots whenever a task demands out-of-distribution generalization or explicit manipulation of abstract variables.

Against this backdrop, the present paper pursues two complementary goals. First, we re-examine how effectively LLMs internalize fine-grained grammatical constraints, an issue frequently probed by comparing the probability distributions assigned by a model to minimal sentence pairs with those elicited from human acceptability judgements \citep{linzen2016assessing,lau2017grammaticality,chowdhuryzamparelli2017rnn,futrell2018rnngrammars,gulordava2018colorless,tran2018importance,wilcox2018fillergap,hu2020systematic,arahalli2024kannada,hosseini2024persian,kim2024ambiguity,ide2025make}. Notably, nearly all prior work centers on English, an abundantly resourced language whose syntactic patterns dominate many training corpora, leaving open how models fare on lower-resourced languages. Second, we ask whether recent algorithmic innovations designed to endow LLMs with explicit multi-step reasoning---exemplified by Chain of Thought style prompting and fine-tuning protocols---can measurably improve a model’s grasp of subtle grammatical phenomena \citep{wei2022chainofthought,wang2024chaindensity,plaat2024treeofthought}.

To that end, we focus on Turkish, a morphologically rich and comparatively under-represented language, and target one particularly informative construction: third-person reflexive anaphors. Turkish supplies two forms, \textit{kendi} and \textit{kendisi}, whose distribution is sensitive to both structural locality and discourse-pragmatic factors. By probing two contrasting systems---OpenAI o1 Mini, advertised as a ``reasoning-oriented'' model, and Trendyol-LLM-7b-base-v0.1, a strong general-purpose Turkish model---we quantify how probability assignments and categorical preferences align with patterns reported in psycholinguistic experiments on human speakers.

\section{Overview of reflexives in Turkish}
Turkish grammar distinguishes between two formally separate but etymologically related third-person reflexive pronouns---\textit{kendi} and \textit{kendisi}---whose distribution has long occupied a central place in the generative literature on binding and anaphoric reference \citep{ozsoy1983kendi, kornfilt2001functional,gurel2003opc,yakut2015perspective}. While both items are glossed as `self', they diverge morphosyntactically: \textit{kendi} is morphologically bare and freely inflects for case (\textit{kendini}, \textit{kendine}, etc.), whereas \textit{kendisi} contains the third-person possessive suffix \textit{-si}, a fact that has been tied to emphatic and logophoric uses. At first glance, this morphological asymmetry suggests a syntactic split: traditional accounts classify \textit{kendi} as a structurally constrained reflexive and \textit{kendisi} as a long-distance or discourse-sensitive anaphor.

Recent experimental investigations, however, complicate this neat dichotomy. Judgement and processing studies by \citet{yakut2015perspective}, \citet{gracanin2017interaction}, \citet{bakaydillon2022ccommand}, and \citet{oguzkaiser2023interpretation} demonstrate that, under certain discourse configurations---particularly those involving perspective-taking or shifted logophoric centers---the two forms converge in acceptability, allowing antecedents that are not clause-internal and thereby contravening the locality requirement suggested in Binding Principle~A \citep{chomsky1981lgb}. These findings raise the possibility that Turkish reflexivization is partly regulated by pragmatic or semantic dimensions (such as viewpoint or topicality) that override purely structural constraints. To ground the discussion, illustrative minimal pairs are provided in examples~(1a) and~(1b) below. In each pair, we juxtapose a structurally local reading (canonically licensed under Principle~A) with a discourse-licensed non-local reading that native speakers nevertheless judge acceptable. This empirical tension makes Turkish reflexives an ideal test bed for probing how language models, and by extension, theories of grammar, integrate syntactic locality with higher-level interpretive factors.

\noindent(1a)\hspace{0.5em}
\begin{tabular}[t]{@{}p{\dimexpr\linewidth-1.2em\relax}@{}}
\raggedright\arraybackslash
Ali$_i$ Fatma-nın kendi-ni izle-diğ-in-i g\"or-d\"u.\\
Ali Fatma-\textsc{gen} self-\textsc{acc} watch-\textsc{nmlz}-3\textsc{sg}-\textsc{acc} see-\textsc{pst}\\
\quad `Ali$_i$ saw that Fatma$_j$ watched self$_{i/j}$.'
\end{tabular}

\vspace{0.35em}

\noindent(1b)\hspace{0.5em}
\begin{tabular}[t]{@{}p{\dimexpr\linewidth-1.2em\relax}@{}}
\raggedright\arraybackslash
Ali$_i$ Fatma-nın kendisi-ni izle-diğ-in-i g\"or-d\"u.\\
Ali Fatma-\textsc{gen} self-\textsc{acc} watch-\textsc{nmlz}-3\textsc{sg}-\textsc{acc} see-\textsc{pst}\\
\quad `Ali$_i$ saw that Fatma$_j$ watched self$_{i/j}$.'
\end{tabular}

Traditional descriptions of Turkish binding treat \textit{kendi} as the prototypically local reflexive: in its canonical use, \textit{kendi} must be bound within the smallest clause (or binding domain) that contains it, conforming to Binding Theory’s Principle~A, which requires an anaphor to have a c-commanding antecedent in a sufficiently local structural domain \citep{georgekornfilt1981finiteness,enc1989binding,kornfilt2001functional,gokselkerslake2005grammar}. In simple matrix clauses and many subordinate environments, this predicts tight clause-boundedness: a matrix subject cannot ordinarily serve as the antecedent for a reflexive embedded in a lower clause, nor can non-c-commanding discourse topics rescue the dependency.

The empirical record is more complicated, however. A parallel strand of work reports that \textit{kendi}, specifically when interpreted with a third-person singular antecedent, sometimes permits reference to an argument located outside its minimal binding domain, yielding what is often termed a long-distance reading \citep{yakut2015perspective}. These exceptions cluster in particular syntactic and discourse configurations: deeply embedded clausal complements of attitude predicates, narrative perspectives that privilege the matrix subject as the ``center of consciousness,'' and contexts where emphasis or contrast appears to license a wider search space for antecedent resolution. Although the precise conditions remain debated and judgements vary across speakers, dialects, and experimental methods, the recurring observation is that structural locality is not an absolute for \textit{kendi} in third-person singular contexts.

In addition to the so-called clause-bound reflexive \textit{kendi}, Turkish grammar furnishes a second reflexive/anaphoric element, \textit{kendisi}, whose referential freedom markedly surpasses the strict locality constraints typically associated with Binding Principle~A.

Morphologically, \textit{kendisi} contains the third-person possessive suffix \textit{-si}, a diachronic clue to its historical development from emphatic or logophoric constructions and to its gradual semantic bleaching into a more general anaphoric form. Syntactically, however, it behaves as a strikingly ``liberated'' anaphor: native-speaker judgements and experimental results \citep{enc1989binding,kornfilt1997turkish} demonstrate that \textit{kendisi} (i) comfortably binds to clause-internal antecedents, replicating the patterns expected for canonical reflexivization, (ii) readily forms dependencies with long-distance antecedents that originate several clauses higher in the structure, and (iii) can even refer to extra-sentential antecedents, discourse-level referents introduced in the broader context but not overtly realized within the sentence containing the anaphor itself.

Taken together, these properties position \textit{kendisi} at the intersection of reflexivity, logophoricity, and discourse anaphora, challenging a strictly locality-based view of binding in Turkish. Its mixed profile also raises important questions for theoretical models that assume a uniform typology of anaphors, suggesting instead that Turkish distinguishes reflexive elements along both structural and discourse-pragmatic dimensions. In this sense, \textit{kendisi} provides an empirical testing ground for evaluating how syntactic, semantic, and pragmatic factors jointly shape anaphoric interpretation.

\section{Large language models as psycholinguistic subjects}

Large-scale language models trained solely on raw, unannotated text have repeatedly surprised linguists by capturing non-local syntactic dependencies that generative theory argues are mediated by hierarchical structure rather than linear proximity. Corpus-based probes and carefully controlled minimal-pair tests show, for example, that current transformers license filler--gap relations, tracking a fronted \textit{wh} phrase and its downstream gap, far above chance \citep{wilcox2018fillergap} and reliably enforce subject--verb number agreement across intervening material in a manner that mirrors human judgements \citep{linzen2016assessing}. These feats are more striking because the models receive no explicit grammatical supervision; the relevant constraints must be abstracted from surface statistics alone. Yet most celebrated successes involve constructions whose forms supply rich, redundant cues: overt agreement morphology, \textit{wh}-fronting markers, or fixed lexical collocations. In these high-signal environments, massive training corpora can permit a model to exploit robust statistical regularities that stand in for deeper grammatical knowledge. Much less is known about how well the same systems cope with dependencies that are rare, structurally opaque, or heavily conditioned by discourse-pragmatic factors, which are situations where surface strings offer few, if any, diagnostic patterns. To probe this gap, a growing literature treats LLMs as surrogate psycholinguistic subjects. Researchers present the network with minimal sentence pairs that differ only at a critical word or phrase whose grammatical status determines acceptability, then ask whether the model assigns higher probability (lower surprisal) to the variant favored by native speakers. Applied to phenomena ranging from agreement to filler--gap licensing, this methodology has produced a mixed record \citep{linzen2016assessing,lau2017grammaticality,futrell2018rnngrammars,gulordava2018colorless,tran2018importance,wilcox2018fillergap}: transformer models excel when the construction is frequent and overtly marked, but their performance deteriorates on low-frequency or syntactically opaque dependencies, casting doubt on whether apparent successes reflect genuine rule induction or sophisticated pattern matching. These findings force the field to confront a fundamental question: What standard should determine linguistic adequacy for artificial language models? From an engineering perspective, the bar is clear: models ought to approximate human behaviour across all constructions, including those rare in training data. From a scientific vantage, however, the aim is to isolate which grammatical regularities are in principle learnable from sequential input, suggesting experimental designs that scaffold exposure and give models a fair chance to display latent competence. Contemporary large language models routinely produce prose that is grammatically well-formed, syntactically diverse, and semantically appropriate, suggesting that they have largely mastered formal linguistic competence, which is the rule-based knowledge underlying natural language structure \citep{mahowald2024dissociating,piantadosi2023modern}. By contrast, their functional competence, which is the ability to deploy general knowledge and reasoning through language, remains contentious \citep{benderkoller2020climbing,marcus2020next,mahowald2024dissociating}.

Although LLMs excel at an impressive array of benchmarks from coherent multi-paragraph generation \citep{brown2020language} to sentiment analysis and logical inference \citep{devlin2019bert,liu2019roberta,radford2019gpt2}, closed-book question answering, theory-of-mind tasks \citep{kosinski2023theory,trott2023evaluating} and selected commonsense-reasoning challenges \citep{zellers2018swag}, closer scrutiny shows that their successes often hinge on shallow word-co-occurrence cues. When such surface regularities are neutralized, model performance frequently collapses \citep{ullman2023llms,she2023bypassing,carranza2025surfaceform,mancoridis2025potemkin}, which is regarded as a fragility that contrasts with the relative robustness of human comprehension \citep{dominguezolmedo2025trainingontesttask}.

\subsection{Chain-of-Thought Reasoning}

One influential response to the brittleness of ``direct-answer'' prompting is to modify the \emph{inference-time} behavior of LLMs rather than their parameters. Chain-of-Thought (CoT) prompting asks the model to produce an explicit sequence of intermediate steps that bridges the question to an answer, effectively turning a single-shot prediction into a multi-step computation \citep{nye2021showyourwork,wei2022chainofthought,kojima2022large}. Beyond simple prompting, related methods (i) sample multiple reasoning traces and aggregate them by voting (\emph{self-consistency}) \citep{wang2022selfconsistency}, (ii) decompose problems into ordered subgoals (\emph{least-to-most prompting}) \citep{zhou2022leasttomost}, (iii) interleave reasoning with external actions such as tool use or information seeking (\emph{ReAct}) \citep{yao2023react}, and (iv) search explicitly over candidate reasoning paths using branching and backtracking (\emph{Tree-of-Thought}) \citep{plaat2024treeofthought}. Closely related proposals further structure intermediate computations as executable programs (\emph{program-of-thought prompting}) \citep{chen2023programofthoughts} or add reflection/self-critique loops to improve deliberation \citep{madaan2023selfrefine,shinn2023reflexion}. Empirically, these strategies often improve accuracy on problems that require compositional arithmetic, multi-hop inference, or constraint satisfaction, and they offer a procedural account of how an output is obtained rather than only what output is produced.

At the same time, it remains unclear whether CoT-style scaffolding strengthens a model’s \emph{linguistic} generalizations in the same way it can strengthen its \emph{task} performance. Grammatical dependencies such as binding are not typically solved by explicit multi-step deduction; instead, they depend on structural locality, hierarchical relations, and discourse constraints that must be represented in the model’s underlying distribution over continuations. This makes Turkish reflexive binding a useful test case: if CoT primarily helps by supporting deliberate search and verification, it may have limited impact on reflexive antecedent preferences; if, however, CoT training induces more systematic sensitivity to structure, it could measurably sharpen locality effects. We therefore contrast a reasoning-oriented model with a Turkish-adapted base model and examine whether CoT correlates with more human-like binding choices. The basic CoT setup is illustrated in (2).

Suppose $Q$ is the question you ask and $A$ is one possible answer. Imagine there is an unobserved chain of thought $C = (c_1,\ldots,c_k)$ that leads from $Q$ to $A$. Chain-of-Thought modeling says that to get the overall probability of $A$ given $Q$, you ``average'' over possible reasoning chains $C$. Formally:
\begin{equation}
\tag{2}
P(A \mid Q) = \sum_{C} P(A \mid Q, C)\, P(C \mid Q)
\end{equation}

\noindent\ Conditional reasoning-chain probability:
\begin{equation}
\tag{3}
P(C \mid Q)=\prod_{i=1}^{k} P(c_i \mid Q, c_{<i})
\end{equation}
This factorization mirrors how large language models generate text: they predict each token (here, each reasoning step) given the accumulated context so far. Because the entire chain is scored---not just a single hop---the model can explore multiple candidate paths and assign higher probability to those whose later steps repair or refine earlier ones.

\noindent\textbf{Reasoning chains:}
\begin{equation}
\tag{4}
\widehat{A}^{\star}=\arg\max_{a}\sum_{j=1}^{N}\mathbf{1}\!\left[\widehat{A}_{j}=a\right]
\end{equation}
Instead of trusting a single chain of thought, the model can generate $N$ different reasoning chains; each chain produces its own answer. We then tally those answers and pick the one that appears most often.

On the other hand, Trendyol-LLM is an open-weight, bilingual Turkish--English language model developed by Trendyol \citep{trendyol2024llm}. It is a decoder-only, auto-regressive transformer based on a LLaMA-family architecture \citep{touvron2023llama2} and fine-tuned using Low-Rank Adaptation \citep{hu2021lora}. As a base model, Trendyol-LLM is not instruction-tuned for dialogue; instead, it is trained to predict the next token given left context. In our experiments we interface with Trendyol-LLM in this standard language-modelling regime: the model processes each stimulus as a single left-to-right sequence and assigns probabilities to the two possible continuations. Thus, while the OpenAI model under study combines large-scale language modelling with an additional reasoning-oriented inference pipeline, Trendyol-LLM instantiates a more ``classical'' setup in which binding preferences must be expressed through the raw next-token distribution. This contrast allows us to ask whether a conventional, Turkish-adapted transformer already exhibits robust locality effects in reflexive binding, and whether reasoning-oriented inference yields qualitatively different behavior on the same materials.

\section{Experiments}

\subsection{The large language models used}

Our experiments juxtapose o1 Mini \citep{openai2024o1} with Trendyol-LLM-7B-base-v0.1 \citep{trendyol2024llm}, built on the LLaMA-2 backbone \citep{touvron2023llama2}, in order to tease apart the contributions of explicit reasoning supervision and Turkish-specific domain adaptation. The o-series models are trained with a reinforcement-learning objective that rewards internally coherent chains of thought. In practice, the learner receives credit not only for producing the correct terminal answer but also for generating intermediate reasoning steps that adhere to global logical consistency constraints. The other model starts from LLaMA-2-7B and undergoes Low-Rank Adaptation (LoRA) \citep{hu2021lora} on \textasciitilde10 billion tokens dominated by Turkish and bilingual web text. Fine-tuning modifies projection matrices in self-attention plus the feed-forward blocks (rank $=64$, $\alpha=128$, dropout $=0.05$).
\subsection{The evaluation metrics}
Large language models have become a workhorse for cognitive modeling because their token-wise probability distributions (and derived measures such as surprisal) can be mapped onto behavioral measures such as reading times and onto neural responses (EEG/fMRI/ECoG), and are also related to graded acceptability \citep{hale2001probabilistic,levy2008expectation,smithlevy2013logarithmic,frank2015erp,lau2017grammaticality,goodkindbicknell2018surprisal,schrimpf2021neuralarchitecture,caucheteuxking2022brainsalgorithms,skrjanecdemberg2026readingtimes}.

. Following this line of work, we evaluate Trendyol-LLM using \emph{sentence-level perplexity} (PPL), defined for a sentence $W = (w_1,\ldots,w_N)$ as:
\setcounter{equation}{4} 
\begin{equation}
\label{eq:ppl}
\mathrm{PPL}(W) = \exp\!\left(-\frac{1}{N}\sum_{i=1}^{N}\log P(w_i \mid w_{<i})\right)
\end{equation}
Sentence-level PPL measures average surprisal (processing cost) per token: lower PPL means the sentence is, on average, more predictable given its context. We illustrate the evaluation using the Turkish stimulus in (6), which mirrors the reflexive configuration introduced earlier.


\noindent(6)\hspace{0.5em}
\begin{tabular}[t]{@{}p{\dimexpr\linewidth-1.2em\relax}@{}}
\raggedright\arraybackslash
\textbf{\label{ex:ppl-stimulus}}%
Ali$_i$ Fatma$_j$-nın kendi-ni izle-diğ-in-i g\"or-d\"u.\\
Ali Fatma-\textsc{gen} self-\textsc{acc} watch-\textsc{nmlz}-3\textsc{sg}-\textsc{acc} see-\textsc{pst}\\
\quad `Ali$_i$ saw that Fatma$_j$ watched self$_{i/j}$.'
\end{tabular}

\vspace{0.6em}

For each item, we generate two minimal continuations that force distinct antecedents for the embedded reflexive. Concretely, we compare a continuation that forces a \emph{matrix-subject} antecedent (7a) to a continuation that forces an \emph{embedded-subject} antecedent (7b).

\vspace{0.35em}

\noindent(7a)\hspace{0.5em}
\begin{tabular}[t]{@{}p{\dimexpr\linewidth-1.2em\relax}@{}}
\raggedright\arraybackslash
\textbf{\label{ex:cont-matrix}}%
Ali izle-di.\\
Ali watch-\textsc{pst}\\
\quad `Ali watched.'
\end{tabular}

\vspace{0.35em}

\noindent(7b)\hspace{0.5em}
\begin{tabular}[t]{@{}p{\dimexpr\linewidth-1.2em\relax}@{}}
\raggedright\arraybackslash
\textbf{\label{ex:cont-embedded}}%
Fatma izle-di.\\
Fatma watch-\textsc{pst}\\
\quad `Fatma watched.'
\end{tabular}

\vspace{0.6em}

Each continuation in (7a) and (7b) instantiates a distinct candidate antecedent for the embedded reflexive in (6). We quantify the model’s binding preference by computing sentence-level perplexity for each continuation: the continuation that attains lower PPL reveals the model’s favored antecedent. Consequently, if continuation (7b) (``Fatma watched.'') yields lower perplexity than continuation (7a) (``Ali watched.''), we infer that the model has resolved the reflexive to \textit{Fatma}. In the case of the o1 model, which does not expose raw probability scores, we instead adopt a forced-choice evaluation, presenting both continuations verbatim and recording which antecedent the model selects.

\subsection{Test set construction}
\label{sec:testset}

The test set consists of 100 carefully constructed Turkish sentences, evenly divided between those containing the reflexive \textit{kendi} (50 items) and those containing the reflexive \textit{kendisi} (50 items). In every sentence, the reflexive is embedded in a nominalized clause, ensuring that two antecedent readings are simultaneously licensed: a local antecedent residing inside the embedded clause and a non-local antecedent in the matrix clause. To manipulate only the reflexive marking while keeping all other lexical material constant, we created minimal pairs, thereby isolating the effect of reflexive type. We further counterbalanced reflexive forms across identical syntactic frames and verbs to eliminate potential frequency and lexical biases. By holding syntactic configuration constant and balancing reflexive type, this design controls for both item frequency and lexical familiarity. Consider the examples in (8).

\vspace{0.35em}

\noindent(8a)\hspace{0.5em}
\begin{tabular}[t]{@{}p{\dimexpr\linewidth-1.2em\relax}@{}}
\raggedright\arraybackslash
\textbf{\label{ex:testset-kendi}}%
Ali$_i$ Fatma$_j$-nın kendi-ni izle-diğ-in-i g\"or-d\"u.\\
Ali Fatma-\textsc{gen} self-\textsc{acc} watch-\textsc{nmlz}-3\textsc{sg}-\textsc{acc} see-\textsc{pst}\\
\quad `Ali$_i$ saw that Fatma$_j$ watched self$_{i/j}$.'
\end{tabular}

\vspace{0.35em}

\noindent(8b)\hspace{0.5em}
\begin{tabular}[t]{@{}p{\dimexpr\linewidth-1.2em\relax}@{}}
\raggedright\arraybackslash
\textbf{\label{ex:testset-kendisi}}%
Ali$_i$ Fatma$_j$-nın kendisi-ni izle-diğ-in-i g\"or-d\"u.\\
Ali Fatma-\textsc{gen} self-\textsc{acc} watch-\textsc{nmlz}-3\textsc{sg}-\textsc{acc} see-\textsc{pst}\\
\quad `Ali$_i$ saw that Fatma$_j$ watched self$_{i/j}$.'
\end{tabular}

\vspace{0.6em}


\section{Results}
\subsection{Evaluation procedure}
\label{sec:eval-proc}

Recall that we evaluated both models on a 100-sentence test set, balanced along one dimension: reflexive type (50 sentences with the bare reflexive \textit{kendi} and 50 with the reflexive \textit{kendisi}). Each sentence embeds the reflexive within a nominalized complement, allowing us to generate two minimal continuations per item: one that forces the reflexive to co-refer with the embedded subject (local reading) and one that forces it to co-refer with the matrix subject (non-local reading). For the open-weight model, we exploited access to raw log-probabilities to compute sentence-level perplexity (PPL) for each continuation.

Crucially, we then reduced this gradient signal to a categorical decision by treating the continuation with lower PPL as the model’s preferred binding. Whenever the embedded-subject continuation yielded lower perplexity than the matrix-subject continuation, we inferred a local preference; conversely, if the matrix-subject continuation yielded lower perplexity, we inferred a non-local preference. This sentence-probability readout approach is standard in work that derives acceptability or preference judgements from language model probabilities \citep{lau2017grammaticality}. For the closed model, by contrast, continuous probability scores are not exposed, so we employed a forced-choice paradigm: both continuations were submitted verbatim, and the model was recorded as selecting one antecedent or the other as its binding decision. In both cases, therefore, the dependent measure is the same---whether the model favoured the local or the non-local antecedent---while the evaluation procedure leverages the most informative interface available for each architecture and avoids conflating performance with superficial output patterns.

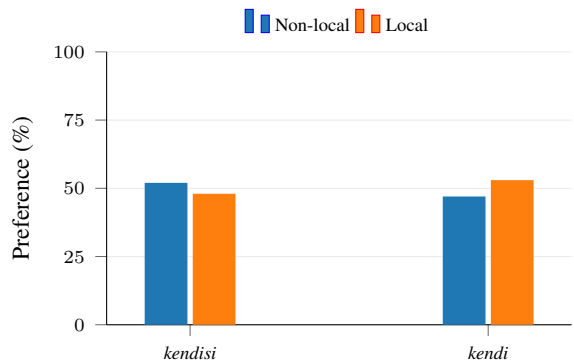
\begin{figure}[t]
\centering
\begin{tikzpicture}
\begin{axis}[
  width=\linewidth,
  height=5.2cm,
  title={o1 Mini},
  ybar,
  ymin=0, ymax=100,
  ylabel={Preference (\%)},
  ytick={0,25,50,75,100},
  ymajorgrids=true,
  major grid style={draw=black!8}, 
  axis lines*=left,
  tick align=outside,
  tick style={black!60},
  label style={font=\small},
  title style={font=\small, yshift=-1.5pt},
  symbolic x coords={kendisi,kendi},
  xtick=data,
  xticklabels={
    \textit{kendisi},
    \textit{kendi}
  },
  xticklabel style={font=\scriptsize},
  yticklabel style={font=\scriptsize},
  bar width=16pt,
  enlarge x limits=0.28,
  legend style={font=\scriptsize, draw=none, at={(0.5,1.02)}, anchor=south},
  legend columns=2,
]
\addplot+[draw=none, fill=mplblue]   coordinates {(kendisi,52.0) (kendi,47.0)};
\addplot+[draw=none, fill=mplorange] coordinates {(kendisi,48.0) (kendi,53.0)};

\legend{Non-local, Local}
\end{axis}
\end{tikzpicture}
\caption{Proportion of trials in which o1-mini favored the non-local vs.\ local antecedent (lower PPL), shown separately for \textit{kendisi} and \textit{kendi}.}
\label{fig:o1mini-binding}
\end{figure}

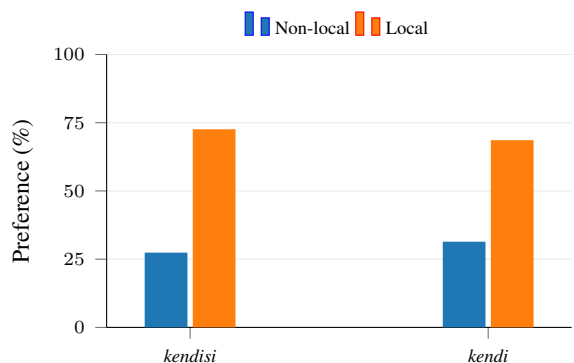
\begin{figure}[t]
\centering
\begin{tikzpicture}
\begin{axis}[
  width=\linewidth,
  height=5.2cm,
  title={Trendyol LLM-7B},
  ybar,
  ymin=0, ymax=100,
  ylabel={Preference (\%)},
  ytick={0,25,50,75,100},
  ymajorgrids=true,
  major grid style={draw=black!8},
  axis lines*=left,
  tick align=outside,
  tick style={black!60},
  label style={font=\small},
  title style={font=\small, yshift=-1.5pt},
  symbolic x coords={kendisi,kendi},
  xtick=data,
  xticklabels={
    \textit{kendisi},
    \textit{kendi}
  },
  xticklabel style={font=\scriptsize},
  yticklabel style={font=\scriptsize},
  bar width=16pt,
  enlarge x limits=0.28,
  legend style={font=\scriptsize, draw=none, at={(0.5,1.02)}, anchor=south},
  legend columns=2,
]
\addplot+[draw=none, fill=mplblue]   coordinates {(kendisi,27.4) (kendi,31.4)};
\addplot+[draw=none, fill=mplorange] coordinates {(kendisi,72.6) (kendi,68.6)};

\legend{Non-local, Local}
\end{axis}
\end{tikzpicture}
\caption{Trendyol-LLM-7B shows a robust locality effect: the local antecedent yields lower PPL in most trials for both \textit{kendisi} and \textit{kendi}.}
\label{fig:trendyol-binding}
\end{figure}

Figure~\ref{fig:o1mini-binding} plots the proportion of trials in which OpenAI o1 Mini favoured the non-local versus local antecedent in both conditions, showing near-chance performance. Figure~\ref{fig:trendyol-binding} shows that Trendyol-LLM-7B, by contrast, exhibits a robust locality effect, with the local antecedent yielding lower PPL in both \textit{kendisi} trials and \textit{kendi} trials.

For Trendyol-LLM, we exploited access to raw token probabilities to compute sentence-level perplexity for each continuation. We then reduced this gradient signal to a categorical binding decision by treating the continuation with lower PPL as the model's preferred antecedent. This probability-based readout is standard in work that derives acceptability or preference judgments from language model probabilities \citep{lau2017grammaticality}. For o1-mini, probability scores are not available \citep{openai2024o1}, so we used the forced-choice paradigm.

\subsection{Statistical analysis}
\label{sec:stats}

A two-tailed binomial test ($H_0 = 0.5$ local preference) confirms that o1 Mini’s behaviour is indistinguishable from chance, whereas Trendyol’s local bias is highly significant ($p < .001$). A $\chi^2$ comparison further shows the two models differ reliably in their antecedent choices. For each sentence the model’s preferred binding is defined as the continuation with lower sentence-level perplexity. Under the null hypothesis of no structural preference ($\pi = .50$ for local antecedents), the number of local choices follows a $\mathrm{Binomial}(N = 100, \pi = .50)$ distribution.

\begin{table*}[t]
  \centering
  \small
  \setlength{\tabcolsep}{3.5pt}
  \renewcommand{\arraystretch}{1.08}
  \begin{tabular}{l c c c p{5.2cm}}
    \hline
    Model & Local choices ($n = 100$) & 95\% BCa CI for $\pi$ & Binomial $p$ & Inference \\
    \hline
    OpenAI o1 Mini  & 48 & .38--.58 & .764  & Indistinguishable from chance \\
    Trendyol-LLM-7B & 71 & .61--.80 & < .001 & Robust locality bias \\
    \hline
  \end{tabular}
  \caption{\label{tab:bootstrap}
    The Bootstrap Analysis. BCa = bias-corrected and accelerated bootstrap, 10,000 resamples
    \citep{efron1987better,efrontibshirani1993bootstrap,diciccioefron1996bootstrap}.
  }
\end{table*}

Table~\ref{tab:bootstrap} summarizes the bootstrap analysis of each model’s tendency to prefer local antecedents. OpenAI o1-mini selected the local antecedent on 48 of 100 trials (95\% BCa CI = .38--.58), a rate that does not differ from the 50\% chance baseline (two-tailed exact binomial test: $p = .76$). By contrast, Trendyol-LLM-7B selected the local antecedent on 71 of 100 trials (95\% BCa CI = .61--.80), indicating a robust locality bias (two-tailed exact binomial test: $p = 3.2 \times 10^{-5}$). Confidence intervals were computed via the bias-corrected and accelerated (BCa) bootstrap with 10,000 resamples, confirming that only Trendyol-LLM-7B shows a statistically reliable preference for local binding. To directly compare the two models’ binding preferences, we constructed a $2 \times 2$ contingency table crossing Model (o1-mini vs.\ Trendyol-LLM-7B) with Binding Choice (Local vs.\ Non-local): o1-mini (48 local, 52 non-local) and Trendyol-LLM-7B (71 local, 29 non-local). A chi-square test of independence shows a reliable association between model type and binding choice:
\[
\chi^2\left(1,\, N=200\right) = 10.98,\; p<.001
\]
\[
\phi = \sqrt{\frac{\chi^2}{N}} = 0.23
\]
Based on the bootstrap results, the $\chi^2$ test of independence (1 degree of freedom; $N = 200$) is significant ($p = .0009$), confirming that the two models differ reliably in their reflexive-binding strategies. The corresponding effect size ($\phi = 0.23$) indicates a small-to-moderate association, suggesting that approximately 5.5\% of the variance in binding choice ($\phi^2 \approx 0.055$) is attributable to model identity (and, by extension, differences in training/adaptation).

\paragraph{Discussion and Conclusion.}
This study set out to discover whether language-specific adaptation or reasoning-oriented supervision is the more decisive factor in aligning large language models with context-sensitive binding behaviour of Turkish third-person reflexive pronouns. To that end, we constructed a 100-item, lexically balanced test set in which the reflexives \textit{kendi} and \textit{kendisi} appear inside embedded clauses, forcing a competition between a structurally local antecedent (the embedded subject) and a clause-external antecedent (the matrix subject). Model preferences were read off from sentence-level perplexity, the exponentiated average negative log-likelihood, which serves as a cognitively interpretable proxy for incremental processing difficulty and has long been used to relate human judgement gradients to language model probabilities \citep{linzen2016assessing,lau2017grammaticality,futrell2018rnngrammars}. For the closed model, we used prompting and the forced-choice paradigm. We compared two contrasting systems: OpenAI o1 Mini, a multilingual model fine-tuned with Chain of Thought supervision that encourages multi-step reasoning \citep{wei2022chainofthought,wang2022selfconsistency}, and Trendyol-LLM-7B, a LLaMA-2 derivative subjected to a LoRA fine-tune on roughly ten billion Turkish tokens drawn from newswire, literature, and e-commerce text. Psycholinguistic evidence shows that Turkish speakers accept both clause-internal and clause-external antecedents, exhibiting at most a graded preference for the embedded subject, an effect that weakens when the matrix subject is thematically demoted or the discourse perspective shifts \citep{bakaydillon2022ccommand,gracanin2017interaction,oguzkaiser2023interpretation}. OpenAI o1 Mini does not show a strong locality bias in this paradigm, which is qualitatively compatible with reports of non-categorical preferences in humans, even though its balance may simply reflect the absence of strong Turkish priors. By contrast, Trendyol-LLM-7B showed an `over-local' bias: in \textasciitilde70\% of cases, the continuation conditioned on the clause-internal antecedent had lower perplexity. The divergence is sizable ($\phi \approx .23$) and unlikely to reflect mere noise: Trendyol is consistently more certain that the embedded subject is the antecedent, whereas o1 Mini distributes its confidence more evenly across local and long-distance readings. At the same time, the precise frequency and distribution of \textit{kendi} and \textit{kendisi} and of the discourse configurations that license long-distance binding in Trendyol-LLM’s training data are not publicly documented, so any account in terms of training-corpus bias must remain necessarily tentative. Against this background, two plausible, but contingent, factors may underlie Trendyol’s stance. First, insofar as the model’s fine-tuning corpus is reported to be dominated by relatively formal Turkish prose, it may underrepresent logophoric or perspectival contexts that facilitate long-distance binding, thereby strengthening structural locality while attenuating discourse-based cues. Second, if clause-internal antecedence is more frequent for these reflexives in the training distribution, the model may have internalized a default local preference that persists even in experimentally balanced materials. We therefore interpret Trendyol’s locality effect as robust within our paradigm, while acknowledging that a definitive attribution to frequency or discourse-distribution effects awaits access to, or detailed statistics from, the training corpus. Furthermore, although Trendyol trims its vocabulary, byte-pair tokenization still fragments agglutinative morphology, and morpheme boundaries often co-occur with embedded-clause boundaries; the model may therefore overgeneralise clause-internal regularities that are artefacts of segmentation \citep{bayram2025tokenization,karakas2026lemmas}. A key limitation is that Trendyol-LLM’s public documentation does not report corpus statistics for \textit{kendi} and \textit{kendisi} or for the discourse contexts that license long-distance binding. Thus, we cannot determine whether the observed locality bias reflects binding constraints or frequency-/exposure-driven preferences.

As a partial external baseline, Aydın’s (2019) corpus study (\textasciitilde450,000 words) reports that \textit{kendi} ranks among the 100 most frequent words in Turkish, suggesting a substantial distributional advantage for \textit{kendi}-type reflexives in naturalistic usage, even though these figures should be interpreted as reference-corpus estimates rather than measurements of the model’s actual training distribution \citep{aydin2019corpus}. These observations cast Chain of Thought in a more circumspect light. CoT makes intermediate inference steps explicit but cannot conjure lexical-syntactic priors absent from the training signal; indeed, it can co-exist with either an over-strong locality effect (Trendyol) or a balanced distribution (o1 Mini) depending on the underlying statistics. Yet, as the model card does not provide token-level statistics for individual lexical items, we are unable to report the training frequencies of \textit{kendi} and \textit{kendisi} for this specific model, and we leave a direct analysis of training-corpus effects to future work, ideally in collaboration with the model developers.

For low-resource languages, this means that reasoning scaffolds must be paired with corpora that faithfully capture both structural and discourse-pragmatic contingencies. Practical levers include morphology-aware tokenizers that respect agglutinative boundaries \citep{bayram2025tokenization,bayram2025tokensmeaninghybridtokenization,lian2025lbpe} and rigorously filtered corpora free from machine-translated noise \citep{cengiz2025evaluating}. Once such foundations are in place, inference-time boosters, multi-round test-time thinking or chain-of-preference optimization, may yield additional gains, but the present findings warn that these techniques will not, by themselves, align models with the graded, context-dependent binding patterns exhibited by humans. Expanding the diagnostic set to discourse-rich stimuli, incorporating newer checkpoints such as OpenAI o4-mini \citep{openai2025o3o4mini} and DeepSeek-R1 \citep{deepseek2025r1}, and testing phenomena where pragmatic cues override syntactic locality will be essential for deciding whether o1 Mini’s balanced profile signals latent grammatical competence or merely random indifference. In sum, genuine human-like linguistic behaviour appears to require a synergistic strategy that marries deep-reasoning objectives to balanced, morphologically informed domain data, rather than privileging either ingredient in isolation.

\bibliography{custom_refs_acl}

\end{document}